\documentclass{article}

\usepackage{arxiv}

\usepackage[utf8]{inputenc} 
\usepackage[T1]{fontenc}    
\usepackage{hyperref}       
\usepackage{url}            
\usepackage{booktabs}       
\usepackage{amsfonts}       
\usepackage{nicefrac}       
\usepackage{microtype}      
\usepackage{lipsum}		
\usepackage{graphicx}
\usepackage[square,sort,comma,numbers]{natbib}
\usepackage{doi}
\usepackage{xcolor}

\title{The Projection-Enhancement Network}

\date{} 					

\author{ \href{https://orcid.org/0000-0003-0878-6792}{\includegraphics[scale=0.06]{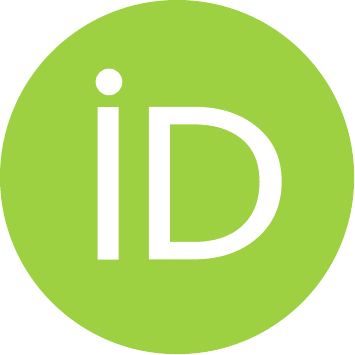}\hspace{1mm}Christopher Z.~Eddy} \\
	Department of Physics\\
	Oregon State University\\
	Corvallis, OR 97331 \\
	\And
	Austin ~Naylor \\
	Department of Physics\\
	Oregon State University\\
	Corvallis, OR 97331 \\
	\And
	\href{https://orcid.org/0000-0001-7001-8781}{\includegraphics[scale=0.06]{orcid.pdf}\hspace{1mm}Bo ~Sun}\thanks{Primary correspondence email: sunb@onid.orst.edu} \\
	Department of Physics\\
	Oregon State University\\
	Corvallis, OR 97331 \\
}

\renewcommand{\shorttitle}{\textit{arXiv} Template}

\hypersetup{
pdftitle={A template for the arxiv style},
pdfsubject={q-bio.NC, q-bio.QM},
pdfauthor={Christopher Z. Eddy},
pdfkeywords={First keyword, Second keyword, More},
}

\begin{document}
\maketitle

\begin{abstract}
	Contemporary approaches to instance segmentation in cell science use 2D or 3D convolutional networks depending on the experiment and data structures. However, limitations in microscopy systems or efforts to prevent phototoxicity commonly require recording sub-optimally sampled data regimes that greatly reduces the utility of such 3D data, especially in crowded environments with significant axial overlap between objects. In such regimes, 2D segmentations are both more reliable for cell morphology and easier to annotate. In this work, we propose the Projection Enhancement Network (PEN), a novel convolutional module which processes the sub-sampled 3D data and produces a 2D RGB semantic compression, and is trained in conjunction with an instance segmentation network of choice to produce 2D segmentations. Our approach combines augmentation to increase cell density using a low-density cell image dataset to train PEN, and curated datasets to evaluate PEN. We show that with PEN, the learned semantic representation in CellPose encodes depth and greatly improves segmentation performance in comparison to maximum intensity projection images as input, but does not similarly aid segmentation in region-based networks like Mask-RCNN. Finally, we dissect the segmentation strength against cell density of PEN with CellPose on disseminated cells from side-by-side spheroids. We present PEN as a data-driven solution to form compressed representations of 3D data that improve 2D segmentations from instance segmentation networks.
\end{abstract}


\section{Introduction}
    Automated computational methods are crucial for high-throughput analyses of microscopy images, where structures of interest are tagged through staining, endogenous expression of fluorophores, or identified through contrast methods. The subsequent image processing, however, often requires expensive expert-level identification \cite{roberts2017CRISPR,ounkomol2018labelfree}. In the domain of cell-science, instance segmentation, or the pixel-wise identification of each unique occurrence of an object in an image, is essential to capture vital morphological and biological insights, and has led to a deeper understanding of cell-heterogeneity \cite{Viana2020HIPSC_PCA}, the spatial-organization of sub-cellular components \cite{gerbin2021Organization,donovan2022Heterogeneity}, and phenotype transitions in cancer \cite{wang2022Epith2Mesen, eddy2021morphodynamics}, to name a few. Deep Neural Networks (DNN) and computer vision methods have been instramental to accomplish these tasks. 
    
    Many biomedical image analyses utilize convolutional neural networks for the identification of objects of interest in their images, in part due to their ability to learn and extract important features in the local receptive fields of stacked convolutions \cite{sarvamangala2021BioMedSurvey,araujo2019receptivefields}. Many of such applications take advantage of two particular architectures, including region-based networks, which propose object regions in an image for downstream segmentation, and U-Net based architectures, which contain an encoder-decoder style network that extracts features and spatial information to construct object segmentations \cite{kar2022benchmarking}.
    
    While many imaging modalities are able to acquire 3D spatial data, several challenges exist in fully-realizing its utility. First, researchers are often limited in 3D resolution due to toxicity or bleaching effects during imaging. To address the issue, computational algorithms have been proposed to infer a high-resolution 3D image from a sub-optimally sampled 3D image stack. The traditional method utilizes deconvolution of the spatially anisotropic point-spread-function with interpolation to overcome the insufficient axial resolution, at the expense of errors in the deconvolution method and additional parameters to hand-tune \cite{dusch2007three, elhayek2011simultaneous}. More recently, state-of-the-art resolution enhancing deep learning techniques have been proposed and proven highly effective for both medical \cite{de2022deep, vaidyanathan2021deep} and microscopy data \cite{weigert2018content, zhang2019high, wang2019deep}. 
    
    When high resolution 3D data is available, it often demands significant overhead in computational time and memory requirements for instance segmentation. Therefore, the majority of current methods do not use an end-to-end approach on 3D data, and instead charge the deep learning networks to only perform semantic segmentation, pixel-wise classification on 2D image slices, and later processed downstream by seeded watershed \cite{fernandez2010MARS, kar2022benchmarking} or other traditional segmentation techniques \cite{wolny2020PlantSeg, wang20223DCellSeg}. When axial resolution is high enough, a different strategy may be to use 2D instance segmentation networks to label a 3D image using all available 2D slices  \cite{stringer2021cellpose}. 
    
    Finally, in training of cell-based DNNs, few public sources of annotated datasets for 3D imaging modalities are available in part due to the tedious nature of annotating such data slice-by-slice. While promising semi-supervised methods have been considered to cut the necessary manual labor costs of annotating data \cite{cciccekSparseAnnotation}, they may introduce unintended bias \cite{chapelle2009semi}. 
    
    
    Due to all these constraints, it is desirable to achieve accurate cell segmentation on 3D image stacks that are sparsely sampled along the axial (z) dimension. The task is particularly challenging at high cell densities. 2D instance segmentation networks have far less parameters involved and offer an end-to-end solution to acquire 2D segmentations of objects. Moreover, there are an abundance of large, readily available labeled 2D cell images through Cell Image Library, Image Data Resource, and Kaggle which can be used to easily train 2D networks. At the single cell level, 2D images also encode most of the morphological quantities that provide accurate phenotype classification \cite{eddy2021morphodynamics}. 

    Given the advantages of utilizing 2D images, it is imperative to recognize the limitations of simple dimensional-reduction approaches. The widely used maximum intensity projection (MIP), for instance, does not provide depth features in order to maintain contextual information, and therefore, spatial context is lost. In segmentation tasks, such as that shown in Figure \ref{fig:fig1}B, MIP introduces spurious overlapping objects that are occluded and results in under-counting and poor segmentation. Other forms of projections, including standard deviation, sum, and mean projections can each introduce their own artifacts into the compressed representation. Instead, many researchers have used color and depth image pairs to overcome the loss of 3D spatial cues to perform 2D instance segmentation \cite{silberman2014instance, gupta2014learning, gupta2015aligning}.
    
    In order to assist cell segmentation in 3D images that are sparsely sampled along the axial dimension, we develop the projection-enhancement network (PEN). PEN is a fully convolutional module which acts as a data-driven unit to encode spatial information of 3D microscopy images into compressed 2D representations. As a module, PEN is placed in front of the 2D instance segmentation network of choice and is trained concurrently to maximize the learning objectives of the instance segmentation network. We show that in contrast to MIP methods, PEN results in significant gains in detection and segmentations in high-density cells in 3D cultures. We show that functionally, PEN learns to encode depth, or becomes a low-high pass filter depending on the training setup. We highlight the segmentation ability of PEN in cancer cells disseminating from spheroids. Considering these results, we present PEN as an effective tool to decrease critical computation time and provide a method to spatially resolve 3D distributed objects in microscopy images for downstream analyses. 

\section{Results}
\label{sec:Results}

\begin{figure}[h]
	    \centering
	    \includegraphics[scale=0.32]{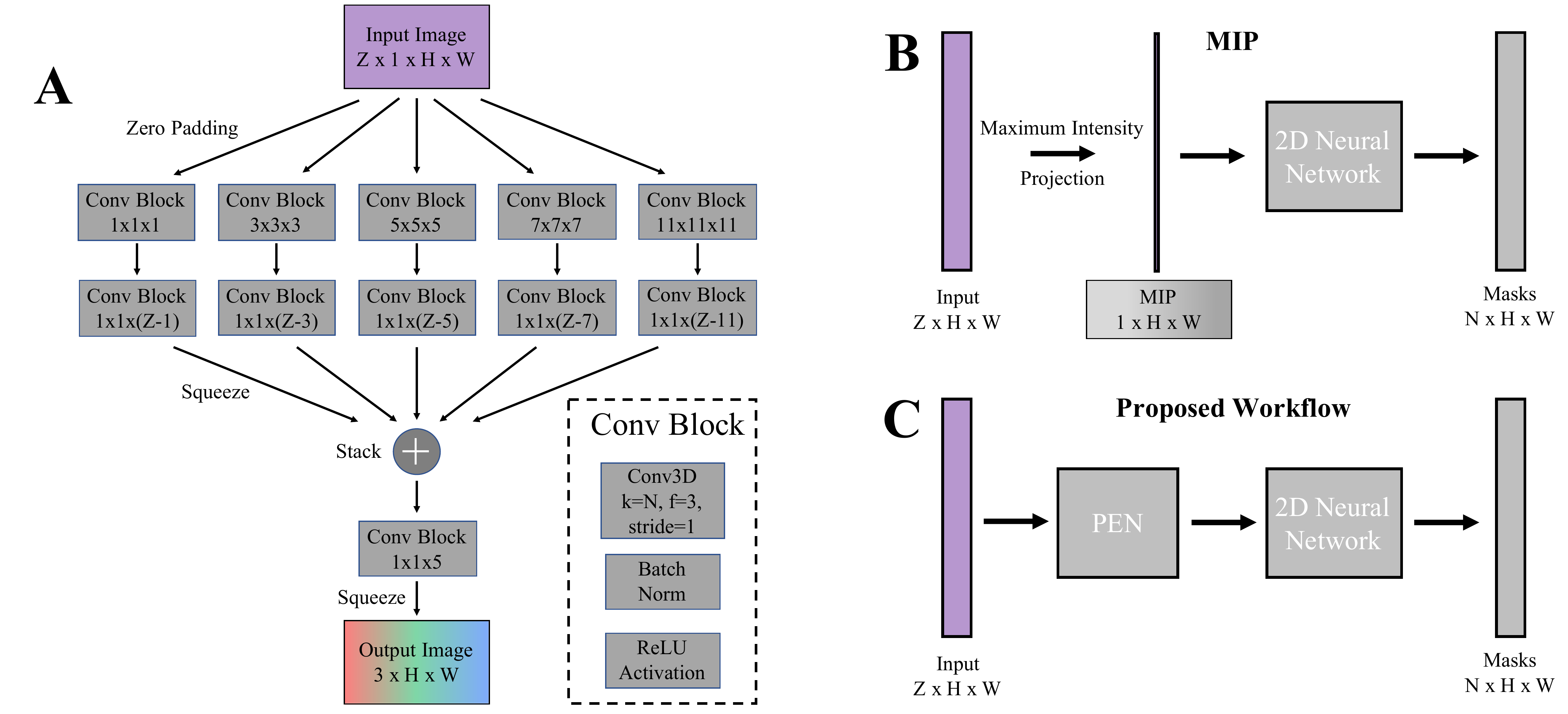}
	    \caption{The Projection-Enhancement Network (PEN). (A) Architecture of PEN to encode 3D axial data into a 2D output image; A Z-stack 3D image is passed as input to PEN, which is operated on by 5 different scales of conv blocks. The outputs of each branch are 3 x H x W, which are then stacked together, and operated on by a final conv block to produce a single RGB image of equal horizontal and vertical resolution as the input. (B) A typical workflow example that used maximum-intensity  projection (MIP) of the input Z-stack for a compressed representation that was passed into a 2D instance segmentation network that predicts object masks. (C) Our proposed workflow diagram of data in the full model. The 3D data is passed to PEN, which passes its 2D RGB output to the 2D instance segmentation network of choice that produces the 2D predicted elements, such as instance masks.}
	    \label{fig:fig1}
    \end{figure}

    In order to take advantage of 2D image segmentation techniques on 3D image stacks that have low axial resolution,  we propose a data-driven model to optimally reduce a gray scale image stack to a 2D RGB image. Our model is inspired by the Inception module \cite{szegedy2015Inception}. A requisite of the model design included forming a shallow network to limit the overhead in terms of memory and computation time, as PEN is built in line with 2D segmentation networks as shown in Figure \ref{fig:fig1}C. Cells distributed in 3D may take any orientation and vary in shape and spatial distribution. These challenges lead us to select an architecture of a wide network, which performs independent operations at multiple scales that are concatenated at the output step. 
    
    Specifically, PEN consists of 3D convolutions distributed in separate branches, as shown in Figure \ref{fig:fig1}A. In each branch, a single convolutional kernel of size K is applied to the 3D image of axial size Z without padding in the axial dimension, and forms 3 feature maps. Following all convolutions, the feature maps undergo ReLU activation, then batch normalization. A subsequent convolution with kernel size of (1, 1, $Z - K$) is applied to pool the axial features. The axial dimension is then squeezed out, and the semantic image in each branch becomes a 2D image with RGB channels. The outputs of the branches are then stacked, and a final 3D convolution is applied with kernel sizes of (1, 1, $N_{branches}$) and 3 output channels. The convolution acts to pool each branch image separately into each output color channel, followed by non-linear ReLU activation and batch-normalization. The third spatial dimension is then squeezed out, leaving a 2D RGB-color image that is rescaled and normalized to be fed to the 2D segmentation network of choice.

\begin{figure}[h]
	    \centering
	    \includegraphics[scale=0.55]{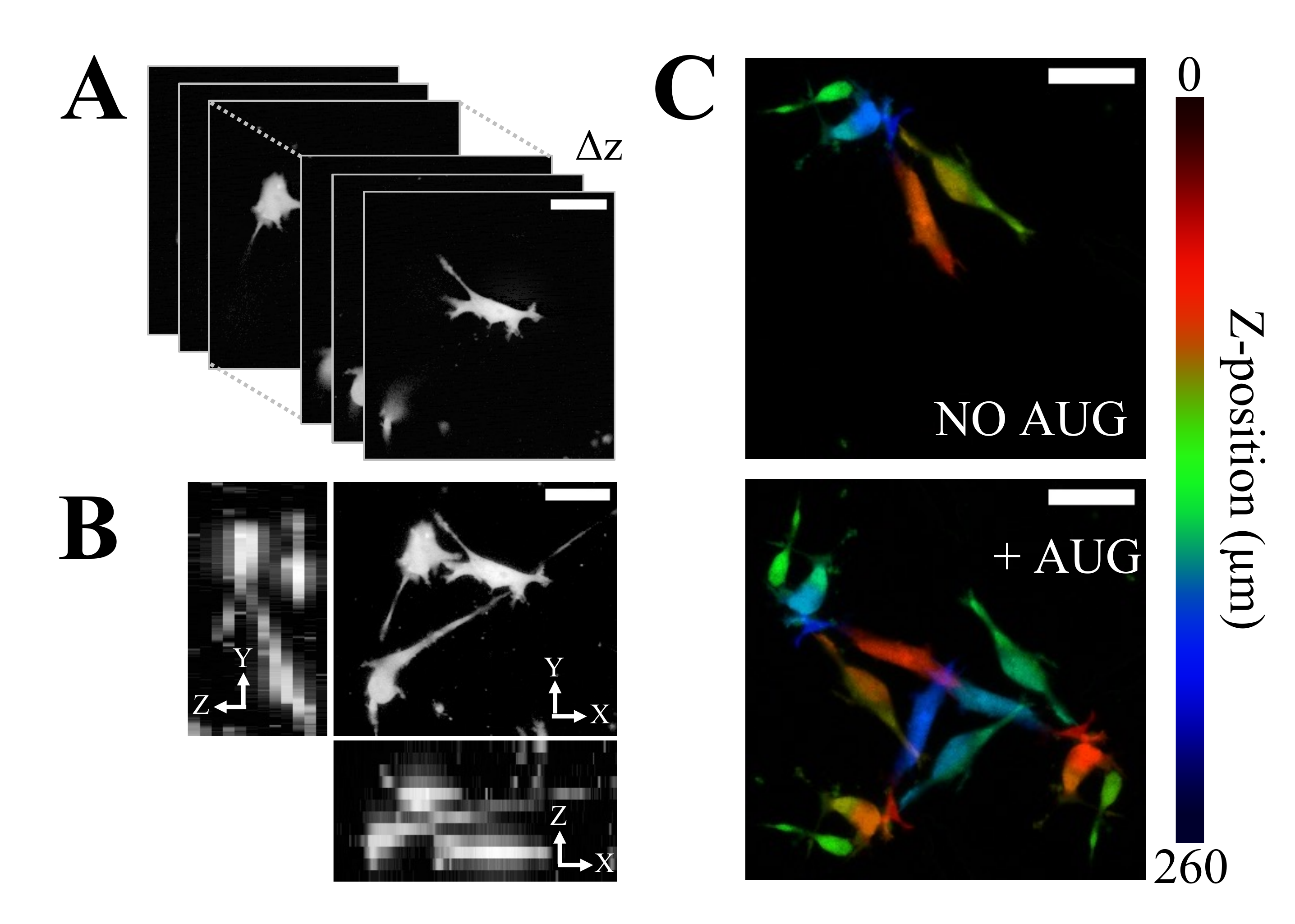}
	    \caption{3D image data for training. (A) MDA-MB-231 GFP cells are embedded in 3D collagen matrices and imaged with confocal microscopy at a low axial resolution of $\Delta$z = 10$\mu$m, resulting in as few as two image slices per cell. (B) MIPs taken over each coordinate axis, where the sub-sampling in the axial dimension results in visible uncertainty in cell boundaries and morphologies in the X- and Y-projections.  (C) Linear depth projections of cell image z-stacks; [top] training image of size 256 x 256 pixels shows 4 cells distributed in 3D with no augmentations applied, and [bottom] same training image with augmentation applied to increase local cell density. See Supplementary S2 for depth projection information. Scale-bars = 30 $\mu$m.}
	    \label{fig:fig2}
    \end{figure}

    Training of DNNs require a large amount of annotated data that has similar characteristics, such as the resolution, cell size and spatial distribution, to the data of interest. Where such data is not available, augmentations can be used to achieve satisfactory performance. The training data utilized here consists of MDA-MB-231 cells recorded with confocal microscopy. As shown in Figure \ref{fig:fig2}A, the gray-scale images are recorded at a low axial resolution of $\Delta$z = 10 $\mu$m, whereas the x-y plane resolution is 0.538 $\mu$m/pixel. As shown in Figure \ref{fig:fig2}B, morphological features of cells are almost completely lost in the axial dimension. The resolution discrepancy associated with the imaging setup, which is often the preferred choice given the photon budget, makes it particularly desirable to perform segmentation based on information in the x-y plane. 
    
    In order to allow segmentation even in datasets of high cell densities, we prepare annotated training images through an augmentation strategy. First, we experimentally obtain confocal images of low cell densities. The confocal stacks cover an axial range of 120 $\mu$m, at steps of 10 $\mu$m, where cells rarely overlap when projected on x-y plane. This makes it easy to annotate cells automatically using simple contrast-based segmentation  and manually correct for errors (Figure \ref{fig:fig2}C top). We then augment the data by artificially duplicating cell images along with their annotations, and apply spatial translation and rotation to combine with the original images. This created an annotated dataset with three times higher cell density (Figure \ref{fig:fig2}C bottom). 
    

\begin{figure}[h]
	    \centering
	    \includegraphics[scale=0.5]{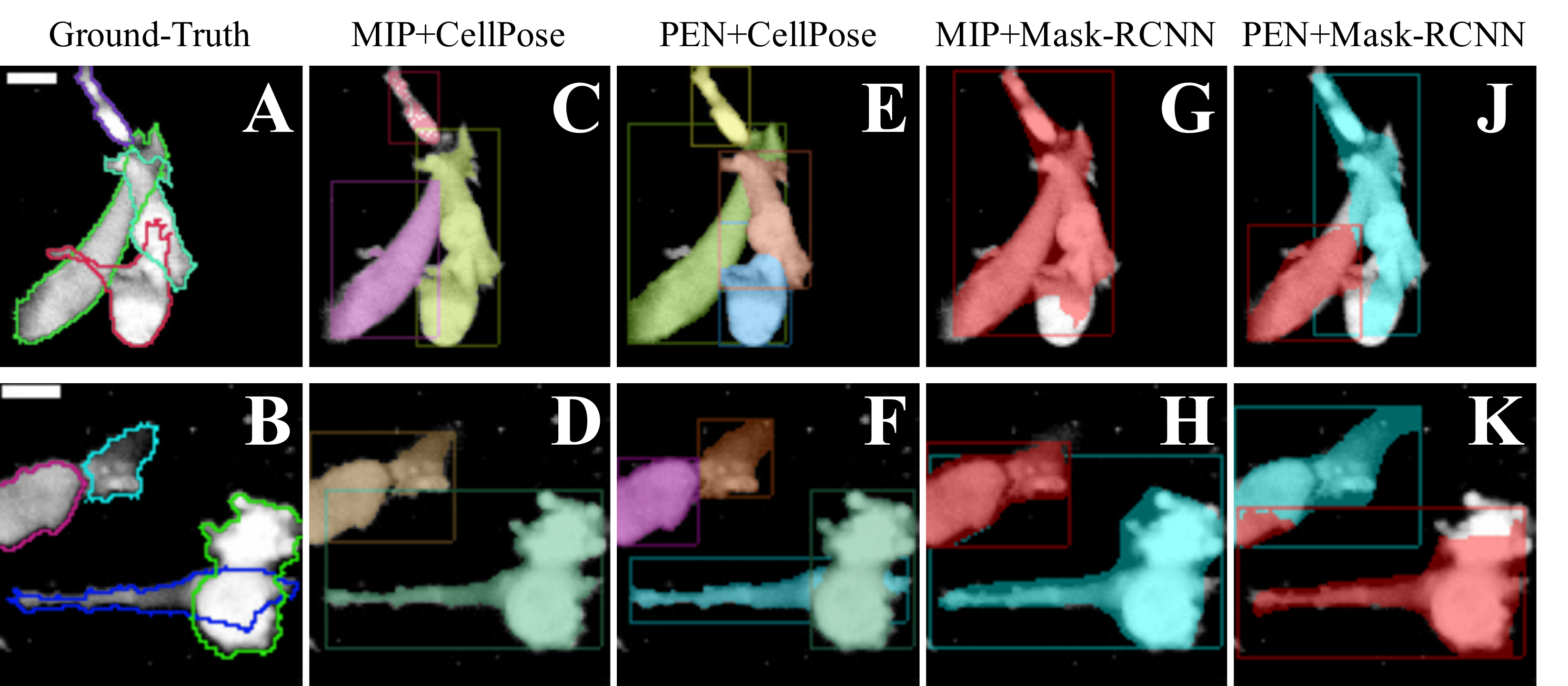}
	    \caption{Comparison of algorithm predictions of cell masks and bounding boxes over two example images. (A-B) Ground-truth object outlines of expert labeled MDA-MB-231 cells are shown in random colors over a MIP image. (C-D) Instance segmentation and bounding box predictions made by MIP+CellPose, (E-F) PEN+CellPose, (G-H) MIP+Mask-RCNN, and (I-J) PEN+Mask-RCNN. Each predicted object is randomly colorized. Scale-bars = 10 $\mu$m.}
	    \label{fig:fig3}
    \end{figure}

    As mentioned in the above, PEN is a module that is placed in front of a 2D segmentation network. We first pair PEN with a modified CellPose network. CellPose is a 2D U-Net architecture that predicts horizontal and vertical flows along with probability maps for cell/background and cell edges for each 2D test image. To resolve multiple cells that overlap when projected on a 2D plane, we modify CellPose to predict $N_{out}$ output channels, where $N_{out}$ is a tunable integer hyperparameter we set to 3 for this work (see also Supplementary S1). For each annotated cell, we assign its label to one of the output channels as ground-truth. The output channel assignment is determined by k-means classification of the axial positions of the annotated cells in the image. Therefore, the depths of cells are monotonically but nonlinearly mapped to the output channels (see also Supplementary S2). As shown in Figure \ref{fig:fig3}C-D, CellPose cannot distinguish cells that are overlapping using maximum-projections as inputs but can correctly identify individual segmentations of cells when PEN is trained in conjunction, as shown in Figure \ref{fig:fig3}E-F.
    

    As a comparison, we also pair PEN with Mask-RCNN (PEN+MaskRCNN). Mask-RCNN is a DNN that consists in part of a Res-Net feature pyramid network which feeds to a RPN that proposes bounding-box regions to later be segmented. Since the RPN may propose overlapping bounding boxes, it may allow for a single 2D pixel to belong to more than one object. We did not modify the output structures of Mask-RCNN, but we expected that the addition of PEN to pass additional 3D spatial information that Mask-RCNN could utilize to distinguish 2D instances. As shown in Figure \ref{fig:fig3}G-J, the addition of PEN does not qualitatively improve the segmentation ability of Mask-RCNN compared to training with MIP inputs. This is consistent with previous reports showing Mask-RCNN often struggles in cases of overlapping instances \cite{suh2021maskrcnn}, as proposed regions in Mask-RCNN during inference are reduced using non-maximum suppression to prevent multiple detections of the same instance.

    To evaluate the performance of different network configurations, we systematically compare four metrics that have been introduced previously \cite{stringer2021cellpose}. The results are shown in Table \ref{tab:table1}. Specifically, we compute the Jaccard Index, Precision, Recall, and a Quality metric which measures the segmentation quality (see also Methods). First, consistent with previous reports, CellPose outperforms Mask-RCNN on the Jaccard Index and has improved segmentation quality \cite{stringer2021cellpose}. Comparing the addition of PEN to each network, on a low-density cell image dataset with > 4,000 annotated cells where fewer than 0.6\% of cells displayed any axial overlap with another cell, the training scheme of CellPose using 2D MIP inputs (MIP+CellPose) slightly outperforms PEN+CellPose on most metrics. However, when compared to a dataset consisting of high-density cell images where 36.8\% of cells had axial overlap with another cell, PEN+CellPose greatly outperforms MIP+CellPose on recall, which measures the ability of the network to detect and segment cells in an image with an intersection over union threshold of 50\%. The poor performance noted in precision is a result of a high frequency of false-positives. Specifically, PEN+CellPose is prone to multiple detections on the same cell, as a result of activation in multiple channels of the output probability maps. On Mask-RCNN, the addition of PEN slightly improves most metrics over both datasets with the consistent exception of the segmentation quality, compared to MIP inputs. However, the performance boost of PEN in Mask-RCNN is less appreciable in comparison to its application in CellPose, particularly in recall on the high-density dataset.
    
    \begin{table}[h]
	    \caption{Quantitative performance instance segmentation networks with the Projection-Enhancement Network (PEN). Performance of CellPose, a U-Net style network, and Mask-RCNN, a region-based network, were evaluated when trained on MIPs or in conjunction with PEN. Models were evaluated on a low-density cell dataset (N = 4082) with fewer than 0.6\% of cells overlapping axially at an average of just 7.2\% intersection over union in the MIP, and a high-density cell dataset (N = 111) where 36.8\% of cells were overlapping axially at an average 12.7\% intersection over union. Metrics are measured at a minimum intersection over union of 50\% for true-positive detections. See Methods for details regarding metrics.}
	    \centering
	    \begin{tabular}{cccccccccc}
        \toprule
        &\multicolumn{4}{c}{Low-Density}
        &
        \multicolumn{4}{c}{High-Density} \\\cmidrule(r){2-5}\cmidrule(l){6-9}
        Model & Jaccard & Precision & Recall & Quality    & Jaccard & Precision & Recall & Quality      \\
        \midrule
        PEN+CellPose & 0.523 & 0.574 & 0.854 & 0.807    & 0.518 & 0.616 & 0.766 & 0.782 \\
        MIP+CellPose & 0.656 & 0.729 & 0.869 & 0.853    & 0.432 & 0.731 & 0.514 & 0.727 \\
        \addlinespace
        PEN+Mask-RCNN & 0.591 & 0.700 & 0.791 & 0.744   & 0.525 & 0.875 & 0.568 & 0.673 \\
        MIP+Mask-RCNN & 0.588 & 0.731 & 0.751 & 0.759   & 0.398 & 0.750 & 0.460 & 0.700 \\
        \bottomrule
    	\end{tabular}
	    \label{tab:table1}
    \end{table}
    
\begin{figure}[h]
	    \centering
	    \includegraphics[scale=0.6]{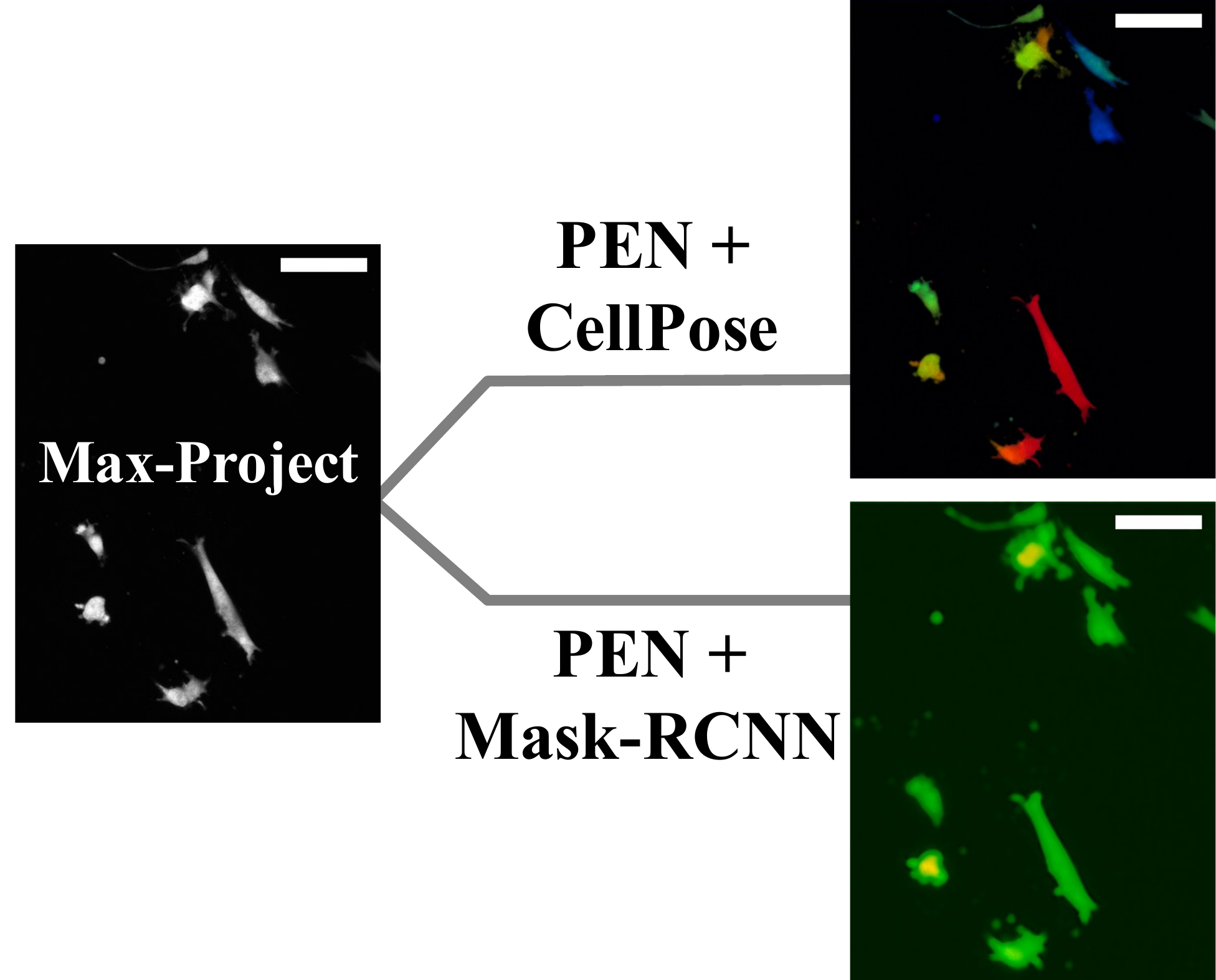}
	    \caption{Outputs of the Projection-Enhancement Network (PEN) after successful training. (Left) A reference MIP image of MDA-MB-231 GFP cells distributed within a 3D image stack, (Top) the output of PEN when trained in conjunction with CellPose, and (Bottom) the output of PEN when trained in conjunction with Mask-RCNN. Scale-bars = 50 $\mu$m.}
	    \label{fig:fig4}
    \end{figure}

    PEN maps a 3D gray scale image stack to a compressed 2D RGB representation, and the mapping algorithm is learned by training PEN in conjunction with a downstream network.  To understand the dependence of PEN on its paired network, we compare the output of PEN when it has been trained with CellPose and Mask-RCNN respectively. Figure \ref{fig:fig4}B shows the output of PEN when trained in conjunction with CellPose, evaluated on a test image stack of low cell density as input. The coloration corresponds to an approximate object depth where cells with lower-to-higher axial positions are mapped to red-to-green-to-blue color channels. This verifies that PEN offers a data-driven approach to color code depth from a 3D image. Inspired by this finding, we further compare PEN to a linear depth embedding algorithm in Supplementary S2 and show that PEN outperforms the linear depth embedding over all metrics analyzed in this work, as seen in Supplementary Table 1. The depth-encoded image from PEN explains why the modules addition to CellPose boosts its performance in the high-density dataset in comparison to training with the max-projection images, as shown in Figure \ref{fig:fig4}A.  Interestingly, we find that when trained with Mask-RCNN, PEN does not similarly encode depth, as shown in Figure \ref{fig:fig4}C. Instead, we find that PEN acts as a low-pass filter, and only maintains activations in one color channel. The blurry low-passed image from PEN demystifies why the quality of segmentations in Mask-RCNN are slightly worse, but makes cells in the image easier to detect and improves precision, recall, and the Jaccard metric.  
    

\begin{figure}[ht!]
	    \centering
	    \includegraphics[scale=0.25]{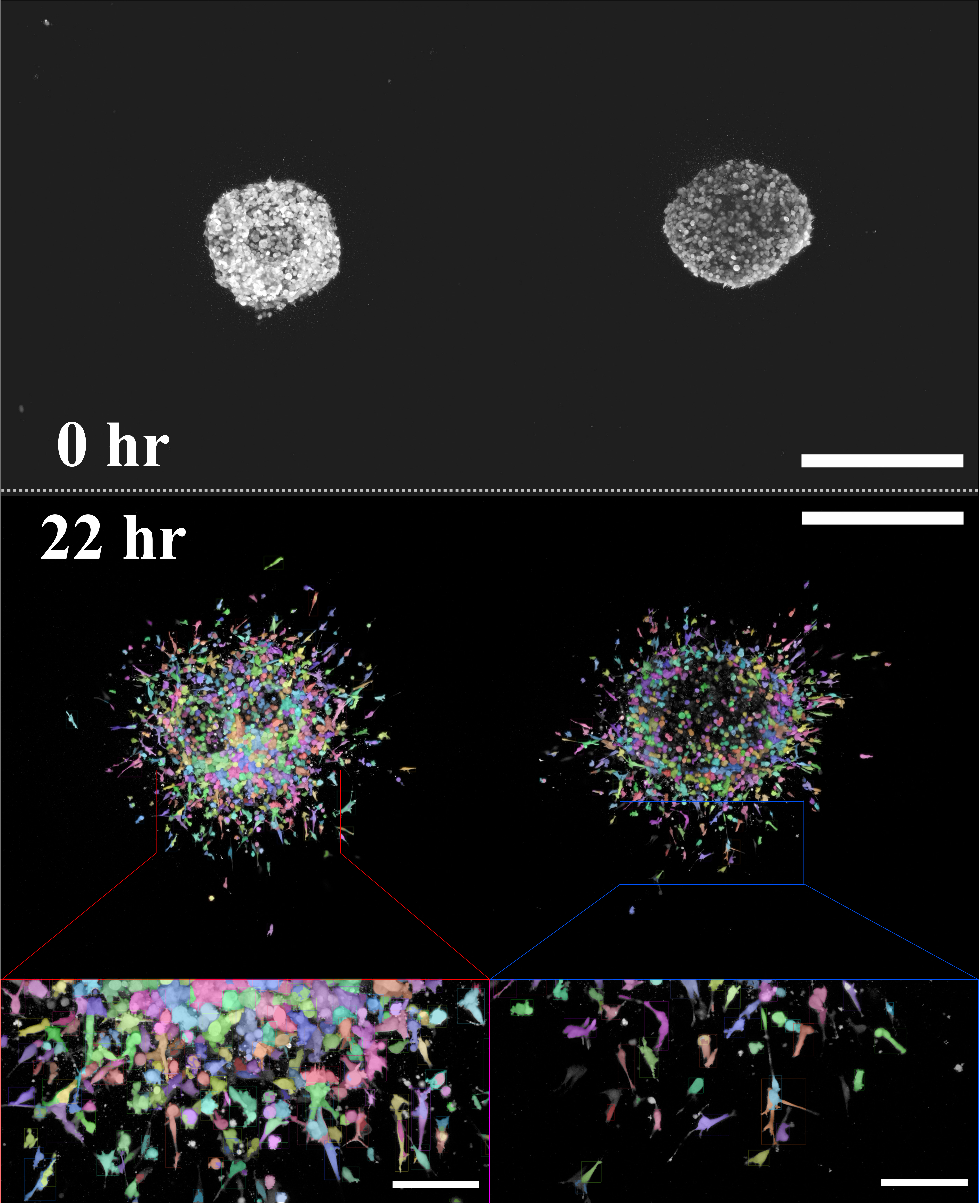}
	    \caption{PEN+CellPose instance segmentation of a 3D system of disseminated cells from side-by-side MDA-MB-231 GFP spheroids. (Top) A MIP image of two MDA-MB-231 GFP spheroids separated by $\approx$1 mm that were gelled in 1.5 mg/mL collagen at 37$^\circ$C and imaged immediately. (Center) The same spheroids were imaged after 22 hours of invasion, and an overlay of the instance segmentation is performed by PEN+CellPose and shown on top of the gray-scale MIP. Over 1700 unique, randomly colored, detections are shown in the lower image. (Lower Insets) Zoomed sections of each spheroid illustrate the effect of crowding on PEN+CellPose performance. Top and Center image scale-bars = 500 $\mu$m, inset scale-bars = 100 $\mu$m.}
	    \label{fig:fig5}
    \end{figure}

    Following successful training with augmented data (Figure \ref{fig:fig2}), we test if PEN+CellPose can handle experimental 3D images with high cell densities and low axial resolution. To this end, we create a sample of two cancer cell spheroids seeded next to each other in 3D collagen matrix (Figure \ref{fig:fig5} top). After 1 day of cell invasion into the matrix, we image the sample with an x-y-z tile scan that covers a volume of 3020 x 1492 x 120 $\mu$m$^3$. The resolution in the x-y plane is 0.538 $\mu$m/pixel, and the resolution in the axial direction is 10 $\mu$m/pixel. Visually (Figure \ref{fig:fig5} bottom), the disseminated cells are identifiable but display significant overlap in the 2D projection. The cells within the spheroid boundary are, however, difficult to distinguish even by an experienced researcher. We apply the trained PEN+CellPose model to the 3D image stack. The segmented cells are randomly colored and plotted over the original data (gray). PEN+CellPose identified 1037 cells associated with the spheroids on the left, and 667 cells associated with the spheroids on the right. Cells disseminated from the spheroids are well segmented. Their elongated shape and various types of protrusions, such as fan-shaped lamellipodia and finger-shaped filopodia, are well preserved. Not surprisingly, the model performs poorly in regions deep within the spheroids. Therefore, we conclude that PEN enables 2D instance segmentation networks to quantify the 3D invasion of tumor spheroids where the imaging covers a large volume under low axial resolution.


    After illustrating the application and performance of PEN for spheroid invasion, we investigate the importance of architectural components to the success of PEN through an ablation study, as shown in Table \ref{tab:table2}. We first examined the contributions of the smallest ($K = 1$) and largest ($K = 11$) convolutional kernel sizes. We find that removal of either kernel does not effect the performance after retraining compared to the base PEN+CellPose model, indicating that the successful axial encoding seen in Figure \ref{fig:fig4}B results from the intermediate kernel sizes, in agreement with the fact that most cells in the training set typically span several slices. However, we expect that including the range of kernel sizes allows PEN to remain robust to new datasets with different axial resolution. 
    
    Next, we investigate replacement of the secondary convolutional block in each branch of PEN with a max-pooling operation over the axial dimension (Branch Max). This alteration makes the network more shallow with fewer parameters to learn. We observed similar performance on the low-density dataset, and a slight decrease in all metrics on the high-density dataset compared to the base model. We speculate that the max-pooling operation makes the network over-reliant on the initial convolution of each branch to learn to incorporate the axial information to the output projection. 
    
    We then explore an alternative method to combine the spatial information learned in each branch of PEN by replacing the final convolutional block with a max-pooling operation (Collect Max). We find that the this model slightly outperforms the base model on recall over both datasets. Here, we choose to keep the convolution despite the comparable performance of the max-pooling layer to maximize the expressive ability of the module, since the pooling operation can be learned by the convolution.
    
    Finally, we investigate the ground-truth assignment strategy used to assign annotated cells to the $N_{out}$ output channels discussed in Supplementary S1. In the base PEN+CellPose model, cells are assigned to $N_{out} = 3$ ground truth output channels based on their z-position, compared to the Random GT model where cells are randomly assigned to  $N_{out} = 3$ channels, and the $N_{out}$ = 1 model where cells are assigned a single output channel. We find that random assignment results in very poor performance across all metrics of both datasets. Additionally, by not including multiple output channels, we increase the performance of the network on the low-density dataset as a result of fewer false-positives, but yields a dramatically decreased performance in recall on the high-density dataset as the network fails to detect superposed objects. We conclude that multiple output channels are vital to the performance of PEN, and that an assignment strategy based on cell position allows PEN to learn and pass axial information to the downstream network.
    
    \begin{table}
	    \caption{Ablation study of the PEN + CellPose (base) instance segmentation network. To evaluate the effects of ablation, each model was retrained from an initialized set of random weights. We evaluated removal of the  $K = 1$ and $K = 11$ kernel sizes, thereby removing an individual branch of PEN shown in Figure \ref{fig:fig1}A. The subsequent convolution in each branch was replaced with a max-pooling operation in the axial dimension in the Branch Max model. The final convolution in PEN was replaced with a max-pooling operation in Collect Max model. Finally, the ground-truth assignment strategy to the available $N_{out}$ channels of our modified CellPose algorithm was set to randomly assign cell labels to $N_{out} = 3$ channels in the Random GT model, and to a single $N_{out} = 1$ channel in the $N_{out}$ = 1 model. All models were evaluated on a low-density cell dataset (N = 4082) with fewer than 0.5\% of cells overlapping axially at an average of just 7.2\% intersection over union in the MIP, and a high-density cell dataset (N = 111) where 36.8\% of cells were overlapping axially at an average 12.7\% intersection over union. Metrics are measured at a minimum intersection over union of 50\% for true-positive detections. See Methods for details regarding metrics.}
	    \centering
	    \begin{tabular}{cccccccccc}
        \toprule
        &\multicolumn{4}{c}{Low-Density}
        &
        \multicolumn{4}{c}{High-Density} \\\cmidrule(r){2-5}\cmidrule(l){6-9}
        Model & Jaccard & Precision & Recall & Quality    & Jaccard & Precision & Recall & Quality      \\
        \midrule
        Base & 0.523 & 0.574 & 0.854 & 0.807            & 0.518 & 0.616 & 0.766 & 0.782 \\
        \addlinespace
        - K = 1 & 0.502 & 0.546 & 0.864 & 8103          & 0.518 & 0.610 & 0.775 & 0.782 \\
        - K = 11 & 0.489 & 0.527 & 0.871 & 0.812      & 0.449 & 0.512 & 0.784 & 0.762 \\
        Branch Max & 0.5202 & 0.564 & 0.870 & 0.818    & 0.4785 & 0.600 & 0.703 & 0.746 \\
        Collect Max & 0.485 & 0.520 & 0.877 & 0.816     & 0.5298 & 0.619 & 0.802 & 0.771 \\
        Random GT & 0.009 & 0.023 & 0.014 & 0.686       & 0.0759 & 0.125 & 0.162 & 0.592 \\
        $N_{out}$ = 1 & 0.6578 & 0.734 & 0.863 & 0.840      & 0.480 & 0.811 & 0.541 & 0.710 \\
        \bottomrule
    	\end{tabular}
	    \label{tab:table2}
    \end{table}
    
\section{Discussion}
\label{sec:Discussion}

    Biomedical research routinely produces 3D image stacks that cover a large volume but have a low axial resolution as limited by practical considerations such as photo damaging, and temporal resolution \cite{jonkman2020tutorial,schneckenburger2021challenges}. To facilitate cell segmentation in such datasets, here we introduce the Project-Enhancement Network (PEN). PEN is a shallow, multiscale, convolutional neural network that encodes a 3D image stack to a 2D RGB color image, which can be subsequently passed to a 2D segmentation algorithm, as shown in Figure \ref{fig:fig1}. We show that when paired with state-of-the-art DNNs of 2D segmentation, PEN enables accurate detection of cells densely populated in 3D image stacks of low axial resolutions, as illustrated in the examples in Figure \ref{fig:fig3}. 
    
    In the training of PEN we take a strategy that leverages data augmentation, which avoids tedious manual labeling to generate annotated data \cite{shorten2019survey}. We find the strategy very effective and can be easily automated by first segmenting low density cell images, then augmenting to artificial high density images, as in the example of Figure \ref{fig:fig2}. Employing this training strategy, we show that PEN+CellPose network can simultaneously detect over one thousand breast cancer cells disseminating from tumor spheroids, as seen in Figure \ref{fig:fig5}. 
    
    We find that the performance of PEN depends on the downstream network it is paired with. In this work, we compared the performance of PEN in conjunction with two leading DNNs in cell-science, CellPose and Mask-RCNN \cite{stringer2021cellpose, matterport_maskrcnn_2017}, as computed in Table \ref{tab:table1}. Significantly, we found that Mask-RCNN did not result in improved performance when built with PEN. A major structural difference in region-based CNNs compared to U-Net style networks is the extraction of regions for segmentation, here through a region proposal network (RPN) in Mask-RCNN. To make the algorithm more efficient, the developers of the RPN in Mask-RCNN chose 3 size-scales and 3 aspect-ratios for the k-anchor boxes proposed within each sliding window \cite{matterport_maskrcnn_2017}. While the network is therefore robust against translations, random orientations and high variance in morphology make many cell-image datasets difficult to determine best size and aspect ratio parameters. In contrast, the efficacy of PEN is purely data-driven and does not restrict object orientation or scale. Furthermore, the RPN has its own loss function to minimize, whereas PEN is only subject to the learning objectives of the instance segmentation network it is attached to and the data that is used as training. On one hand, no additional loss function is a feature of PEN, making it light-weight and a plug-and-play module. However, on the other hand, no direct learning objective makes PEN susceptible to learn inconsistent or poor feature embeddings as a result of underlying patterns in the data. Taken all together, we suggest PEN to be paired with non-region-based downstream networks, specifically U-Net style segmentation networks.
    
    Our results shed light on the explainability of DNNs \cite{tjoa2020survey}, as visualized in Figure \ref{fig:fig4}. In the PEN+CellPose configuration, we show that after training, PEN learns to become a nonlinear depth encoder. This makes it possible for the 2D CellPose to detect overlapping cells on a 2D plane using the depth-encoding color information. In the PEN+MaskRCNN configration, however, PEN learns to become a low-pass filter. We speculate that the non-maximum suppression used in region-proposal networks to filter out multiple detections of objects with significant intersection-over-union prevents Mask-RCNN from detecting overlapping cells in any 2D projected image. However, the learned embedding helps to improve the segmentation of the single detected object, as edges are more easily distinguished in the low-pass image. Therefore, we find that after training, PEN turns an input image into a semantic embedding that represents the best image transformation to maximize the learning objectives of the neural network it is attached to. 
    
    Through a systematic ablation study in Table \ref{tab:table2}, we find the performance of PEN+CellPose critically depends on the assignment strategy of ground-truth annotations to multiple $N_{out}$ predicted channels.  The modifications of CellPose in this work, particularly expanding the predicted maps to multiple channels corresponding to object depth in the 3D image stack, are vital to detect overlapping cells in 3D. Indeed, reducing $N_{out}$ from 3 to 1 seriously deteriorates the segmentation performance. It is interesting for future studies to further explore the optimal $N_{out}$ that balance the computational cost and segmentation power.

    In conclusion, we propose PEN as a plug-and-play module that provides a data-driven approach to compress a 3D image stack into a 2D RGB representation as inputs for 2D instance segmentation networks. We highlight PEN's utility in the detection of disseminated cells from cell-dense spheroids and in settings of significant cell-cell overlap. Our result is a deep-learning solution for instance segmentation in a data regime often overlooked in the field. We envision PEN to be a useful tool for a wide range of applications such as in research of cancer and developmental biology.

\section{Methods}
\label{sec:Methods}

\subsection{Maintenance of MDA-MB-231 GFP Cells}
\label{subsec:Methods/MDACulture}
    GFP-labeled MBA-MB-231 human breast carcinoma cells are purchased from GenTarget Inc. and are maintained according to the manufacturer's instructions. Briefly, growth media is prepared using Dulbecco's Modified Eagle Medium (Gibco, US) supplemented with 10\% fetal bovine serum (Gibco, US), 1\% penicillinstreptomyocin (Gibco, US), and 0.1 mM non-essential amino acid (NEAA 100x, ThermoFisher, US). Generally, cells are cultured at less than 80\% confluency and seeded on culture dishes at recommended concentrations and maintained for up to 12 passages. Cells are kept in culture flasks in a tissue culture incubator at 37$^\circ$C and 5\% CO$_2$.

\subsection{3D Cell Culture}
\label{subsec:Methods/3DCellCulture}
    Training images were acquired from experiments of GFP-labeled MDA-MB-231 cells dispersed in 3D collagen matrices at a low cell-density. For these experiments, collagen solutions were prepared by diluting rat-tail collagen type I (Corning, US) with prepared growth medium, phosphate-buffered saline (PBS, 10x), and sodium hydroxide (NaOH, 0.1M) to a concentration of 1.5 mg/mL or 3.0 mg/mL with pH 7.4. To embed the cells in 3D collagen matrices, cells are suspended at very low density of approximately 650 cells/$\mu$L in ice-cold neutralized collagen solution and added to a 35 mm collagen coated glass bottom dish with a 7 mm microwell diameter (No. 0 coverslip, MatTek, US). The microwell containing ice-cold cell-collagen solution is covered with a coverslip so that the dish may be inverted during gelation to ensure dispersion of cells in 3D. The dish is then incubated on either a warming plate set to 25$^\circ$C, or in a tissue culture incubator (37$^\circ$C, 5\% CO$_2$) for 30 minutes in order to solidify the matrix. The coverslip is removed after gelation time and the cellularized ECM is immersed with tissue culture medium and continuously incubated for 24 hours before imaging. Prior to imaging, 1 M HEPES (Gibco, U.S.) is added to 10\% v/v to DMEM growth media in microwell MatTek dishes containing cellularized ECM to maintain pH during imaging. The dish is then imaged as described in Methods. 

\subsection{3D Spheroid Culture}
    GFP-labeled MDA-MB-231 spheroids seen in Figure \ref{fig:fig5} are grown following methods by Thermo Fisher Scientific \cite{ThermoFisher_Spheroid}. Briefly, MDA-MB-231 cells are first cultured and seeded at low density in growth medium (100 cells/$\mu$L) in a 96-well low-attachment dish (manufacturer). Cells are collected at the bottom of the dish by centrifuging at 290 g for 3 minutes. After overnight culture in a tissue culture incubator (37$^\circ$C, 5\% CO$_2$), rat-tail collagen type I (Corning, US) is added to each well to a final concentration of 6 $\mu$g/mL in order to promote compact spheroid growth. The 96-well plate is again centrifuged at 100 g for 3 minutes, and placed back into the tissue culture incubator. Between 3-5 days later, spheroids are cultured as follows. First, an ice-cold neutralized collagen solution is prepared as previously described to a collagen concentration of 1.5 mg/mL and final pH of 7.4 and kept on ice. Spheroids are then detached from the 96 well-plate by gentle expulsion of growth media with a pipette. Once the spheroid is visibly free-floating, the spheroid is gently pipetted out of the well and expelled into the ice-cold collagen solution. Multiple spheroids may be added to the same collagen solution as desired. The spheroid-collagen solution is then added to 35 mm collagen coated glass bottom dish with a 7 mm microwell diameter (No. 0 coverslip, MatTek, US). The dish is then placed into the tissue-culture incubator for 15 minutes to solidify the matrix, then removed and immersed in DMEM culture medium with 1M HEPES (Gibco, U.S.) added to 10\% v/v. As invasion proceeds rapidly (within hours), the dish is immediately taken for imaging.

\subsection{Microscopy}
\label{subsec:Methods/Microscopy}
    3D imaging is done with a Leica TCS SPE confocal microscope with a 20x oil immersion lens (NA 0.60) equipped with a stage-top incubator (Ibidi). Generally, experiment dishes are placed on an on-stage incubator (Ibidi) which maintains a constant 37$^\circ$C temperature during imaging.  A drop of the immersion oil (type HF, Carquille, U.S.) is left in contact between the dish and the objective lens to equilibrate for an additional half hour to prevent drift while imaging. The acquired raw images are gray-scale with a resolution of 1024 x 1024 pixel$^2$. The voxel size has been calibrated to equal 0.538 $\mu$m. A single x-y plane is imaged every 10 $\mu$m in the z-dimension per experiment, resulting in as few as 2 images per cell depending on the orientation and morphology of the cell.
    
\subsection{Quantitative Metrics}
\label{subsec:Methods/Metrics}
    In order to make direct comparisons, we have used the same analysis used in the original CellPose work to analyze the performance of CellPose and Mask-RCNN networks with and without the addition of PEN \cite{stringer2021cellpose}. Briefly, predicted objects are assigned to ground-truth labels, and thus labeled as true-positives, using a linear sum assignment to minimize the intersection over union loss. However, we require predicted objects to have an intersection over union with their corresponding ground-truth assignment of 0.5 to be an eligible candidate for a true-positive (TP) label. All non-matched predictions are labeled as false-positives (FP), while all missed ground-truth objects are labeled as false-negatives (FN).  In this work, the metric of Jaccard Index is the same definition as average precision used in the original CellPose work, defined as 
    \begin{equation}
        Jaccard\: Index = \frac{TP}{TP + FP + FN}. 
    \end{equation}
    Precision is used to measure the percentage of false-positive predictions made by the deep-learning segmentation, and defined as 
    \begin{equation}
        Precision = \frac{TP}{TP + FP}. 
    \end{equation}
    Recall measures the ability of the network to detect and segment objects in an image, and is defined as 
    \begin{equation}
        Recall = \frac{TP}{TP + FN}. 
    \end{equation}
    Finally, because the definition of Precision here does not measure the ratio of properly identified pixels to all predicted object pixels, we include a final metric we call "Quality". This metric measures the average segmentation quality, and is defined as the average intersection over union of true-positive elements, or
    \begin{equation}
        Quality = \overline{IoU}_{matched}. 
    \end{equation}

\subsection{Dataset Information}
\label{subsec:Methods/Dataset}
    For the work reviewed here, the acquired images have not undergone any additional image processing – a testament to the effectiveness of the deep-learning networks to detect cells. In order to process z-stacks, image planes are stored within OME-TIFF file formats using Tifffile Python library. Ground-truth annotations were acquired by manual thresholding of z-stacks and taking MIPs of individual cells. Specifically, prior to thresholding, fluorescence images are background subtracted using a rolling ball radius of 50 pixels (26.88 $\mu$m) and then log-transformed in order to make cell edges highly visible and so that less fluorescent-intense cells are also quantified using ImageJ (NIH). A manual threshold is then applied for each image. After, cells are manually segmented for each z-stack if applicable. Since consecutive z-stacks may have cell overlap, custom Matlab scripts are then used to determine if the same cell is in multiple z-stacks. After, we take a MIP (2D) of each cell. We then save the cell masks as sets of vertices using scikit-image for a compact representation of the data in a JSON format. The ground-truth vertices are then imported and converted back to masks and cell borders, and horizontal and vertical gradients are calculated for the CellPose algorithm using the heating algorithm described by \cite{stringer2021cellpose}. A subset of the training and validation data, the curated high cell density data, and the spheroid image of Figure \ref{fig:fig5} is shared on Figshare at \cite{naylor_2022_train,naylor_2022_curated,eddy_2022_Z_image}. The full training set is available upon request.

\subsection{Network Training Procedures}
\label{subsec:Methods/Training}
    All CellPose networks with and without PEN described in this work were trained on a single Nvidia Tesla K80 12gB GPU. Training images were 256 x 256 pixel$^2$ in size, and a batch size set to 8 images, and networks were trained for 50 epochs with 50 iterations per epoch. The results discussed in this work were taken from the model weights minimizing the total loss of a 100 image validation set, which underwent the same augmentation and cropping prescribed for the training set. The total loss function for CellPose networks was a summation of cross-entropy loss for the cell-background probability map, a mean-squared error loss for the predicted horizontal and vertical gradients, and a dice-loss for the probability map of cell edges. Training used stochastic gradient descent with momentum set to 0.9, with a learning rate of 0.02, a weight decay of 1e$^-5$ for regularization, and gradient clipping of 5 to prevent exploding gradients during.
    
    All Mask-RCNN networks are similarly trained on a single Nvidia Tesla K80 12gB GPU. Training images are 512 x 512 pixel$^2$ in size, and a batch size set to 2 images, and networks were trained for 50 epochs with 50 iterations per epoch. We use the same weighting scheme for losses and follow the same training procedure as used in the original implementation by \cite{matterport_maskrcnn_2017}. We use a ResNet-50 backbone, RPN anchor scales of 8, 32, 64, 128, and 256, and anchor ratios of 0.5, 1, and 2. We use an NMS threshold of 0.9 during training, and reduce the threshold to 0.7 during inference to consider more proposed regions.

\subsection{Code Availability}
    The PEN was developed to be a simple plug-and-play module, easily implemented on top of any 2D instance segmentation network that accepts the typical 2D RGB input image structure. We are in the processing of developing PEN into an installable Python library through the Python Package Index. CellPose was developed by Stringer and Pachitariu and originally written in PyTorch \cite{stringer2021cellpose}. We have translated the open-source code to Tensorflow/Keras and have made several modifications, such as prediction of cell-edges, the multichannel output discussed in Supplementary S1, and a faster post-processing flow algorithm, but we make no claim on their intellectual property. Mask-RCNN was developed by He and Girshik and the implementation developed with Tensorflow/Keras by Abdullah \cite{he2017maskrcnn, matterport_maskrcnn_2017}. We have only modified the configuration files to import the data structures discussed in this work, and to build PEN on top of their 2D network. Details of training can be found in Methods. All source code developed in this work and trained models are available at https://github.com/eddy6081/PEN.

\bibliographystyle{abbrv}
\bibliography{Main}  






\end{document}


\maketitle

\section{Supplementary}
\label{sec:Supplementary}

\subsection{CellPose Network Modifications}
\label{subsec:Supp/CellPoseMods}
    The original 2D CellPose network outputs 3 predicted elements, including a cell/background probability map, and the horizontal and vertical gradient predictions, with just a single slice (channel) for each predicted element corresponding to the single 2D image used as input. In this work, since we seek to incorporate the 3D nature of the image stacks to help segment overlapping cells, a single output slice would not allow the network to identify overlapping cells. Instead, we add a single parameter $N_{out}$ that must be assigned to decide number of ground-truth slices for each predicted element. We use this parameter to assign cells in the image stack to $N_{out}$ K-means clusters. This assignment and downstream predictions can be analogously considered to breaking the image stack up into $N_{out}$ sub-stacks over which MIPs are taken. In this work, $N_{out}$ is default set to 3. In the case where the large H x W image contains fewer than or equal to $N_{out}$ cells, the cells in the image are assigned to the available channels subsequently based on axial position. Otherwise, the cells are assigned following a K-means assignment over all the cell axial positions in the H x W image. Regarding details of the K-means analysis, the K-means initial cluster positions are linearly equidistant based on the image stack size in order to be reproducible between multiple epochs of training. The K-means algorithm is set to run for 300 iterations. After, augmentation of the training image stack and ground-truth masks are applied, cell edges and horizontal and vertical gradients are calculated as discussed in Supplementary \ref{subsec:Supp/DataDetails}.

\subsection{Comparison of PEN to Linear Depth Embedding Algorithm}
\label{subsec:Supp/PENvsLinear}


    \begin{figure}[h]
	    \centering
	    \includegraphics[scale=0.6]{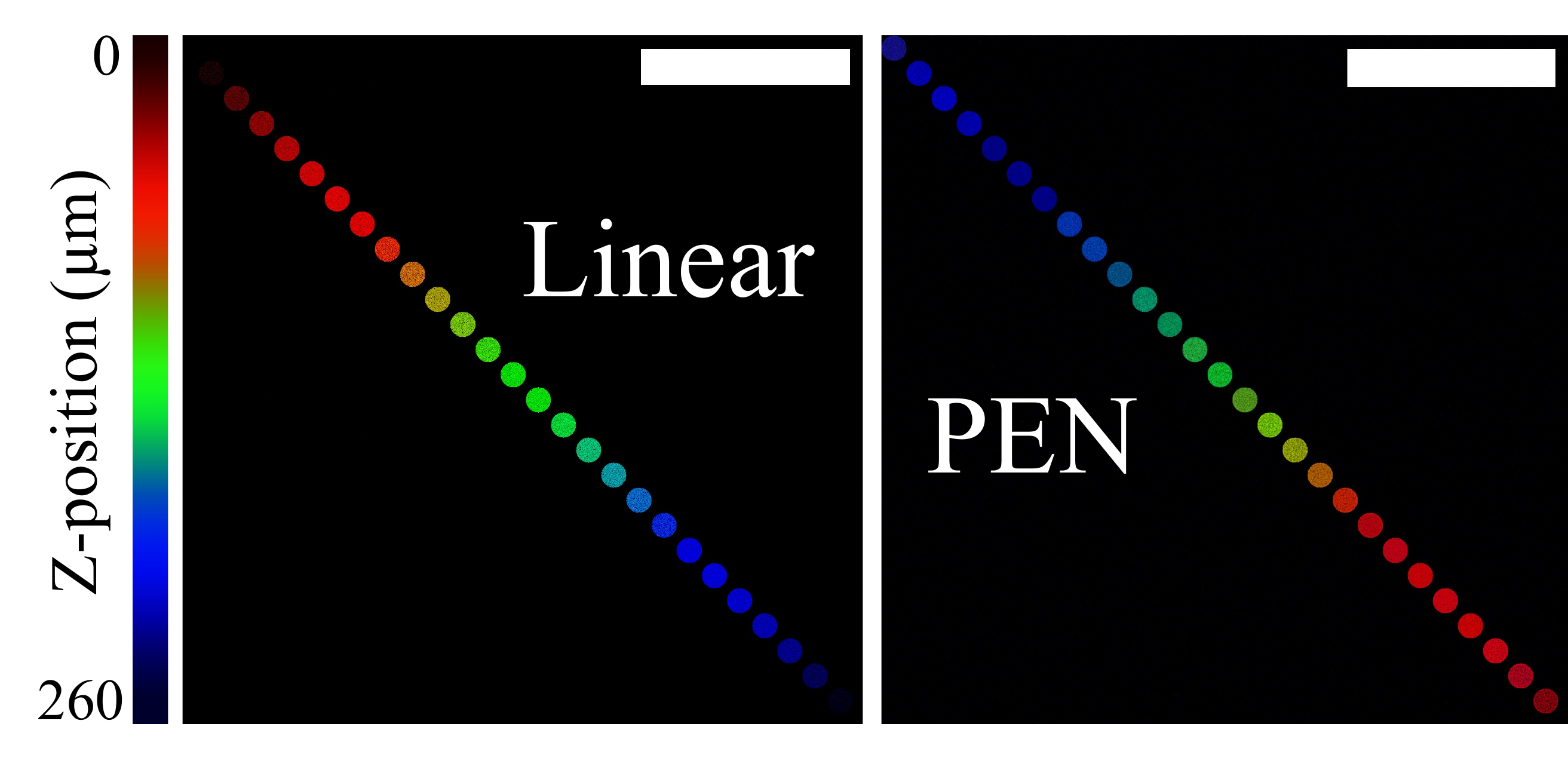}
	    \caption{Comparison of a linear depth embedding algorithm to PEN. Disk images are computationally generated so that a single disk of diameter 30 $\mu$m is placed in an individual slice of the image, and translated along the diagonal between subsequent slices. (Left) Z-stack disk image is linearly depth embedded and compressed to a RGB image. (Right) Output of PEN after processing Z-stack image, following successful training of PEN in conjunction with CellPose. Color-bar for linear embedding algorithm is shown to the left. Scale-bars = 250 $\mu$m.}
	    \label{fig:sup2}
    \end{figure}

    As shown in Figure 4 of the main text, we observed that the output of PEN when trained in conjunction with CellPose seemed to be indicative of the axial position in the image stack. In order to investigate if PEN encodes depth, we created an artificial 3D image containing disks with similar diameter to those of cells in our images. Each slice of the 3D stack contains a single disk of diameter 30 $\mu$m. The disk is then translated along the diagonal a diameter amount horizontally and vertically and placed in the next subsequent slice. 
    
    We next linearly depth embed the image as shown in Figure \ref{fig:sup2}. The linear depth embedding algorithm calculates each output channel by multiplying each z-slice image by a corresponding point in a normal distribution for each output channel. The normal distributions are centered so they are linearly separated between the number of slices of the 3D image, and the standard deviation of each curve set so that the FWHM is at 50\% between the peaks of each normal distribution.
    
    We next evaluated the output of PEN when passed the 3D disk image as input, after training PEN in conjunction with CellPose. As shown in Figure \ref{fig:sup2}, comparing the linear depth embedded image to PEN, we find that PEN similarly learned to encode depth. Furthermore, we compared the performance of the CellPose network when trained using a linear depth embedding algorithm for 3D input images, instead of PEN. As shown in Table \ref{tab:supp_tab}, We find that PEN outperforms the linear depth embedding algorithm across all metrics using a high cell density dataset with significant axial overlap between cells. We conclude that the current architecture of PEN learns a data-driven depth embedding that is an improvement over a simple linear depth embedding algorithm.

    \begin{table}[h]
	    \caption{Performance comparison of CellPose trained in conjunction with CellPose or trained using linearly depth embedded images as input. Both models were evaluated on a high-density cell dataset (N = 111) where 36.8\% of cells were overlapping axially at an average 12.7\% intersection over union. Metrics are measured at a minimum intersection over union of 50\% for true-positive detections. See Methods section in the main text for details regarding metrics.}
	    \centering
	    \begin{tabular}{ccccc}
        \toprule
        &\multicolumn{4}{c}{High-Density} 
        \\\cmidrule(r){2-5}
        Model & Jaccard & Precision & Recall & Quality \\
        \midrule
        PEN+CellPose & 0.523 & 0.574 & 0.854 & 0.807 \\
        Linear+CellPose & 0.446 & 0.549 & 0.703 & 0.758 \\
        \bottomrule
    	\end{tabular}
	    \label{tab:supp_tab}
    \end{table}

\subsection{Details of Datasets and DNN Training}
\label{subsec:Supp/DataDetails}
    To promote PEN’s ability to distinguish overlapping cells, we rely heavily on augmentation during training. First, the mean value of the image stack is subtracted from each pixel. Then, the image stack and corresponding set of 2D masks are centered on a cell centroid and then cropped to the training height and width. The image stack and cell masks are then copied, and the copied stacks are randomly rotated and flipped in each spatial dimension, and then the copied image stack is translated a random integer value by padding at the front of the axial dimension. The original image stack is padded the same amount at the end in the axial dimension, and the copied image and original image are combined pixel-wise such that the maximum value is taken. The combined image is then center padded as necessary to the fixed input axial dimension size (27 slices). The copied mask stack and original stack are concatenated in the axial dimension with size equal to the number of ground truth labels after augmentation. Finally, the recombined image stack and final stack of cell masks are again randomly rotated and flipped in each spatial dimension to assure the network becomes rotation invariant. 
    
\begin{figure}[h]
	    \centering
	    \includegraphics[scale=0.52]{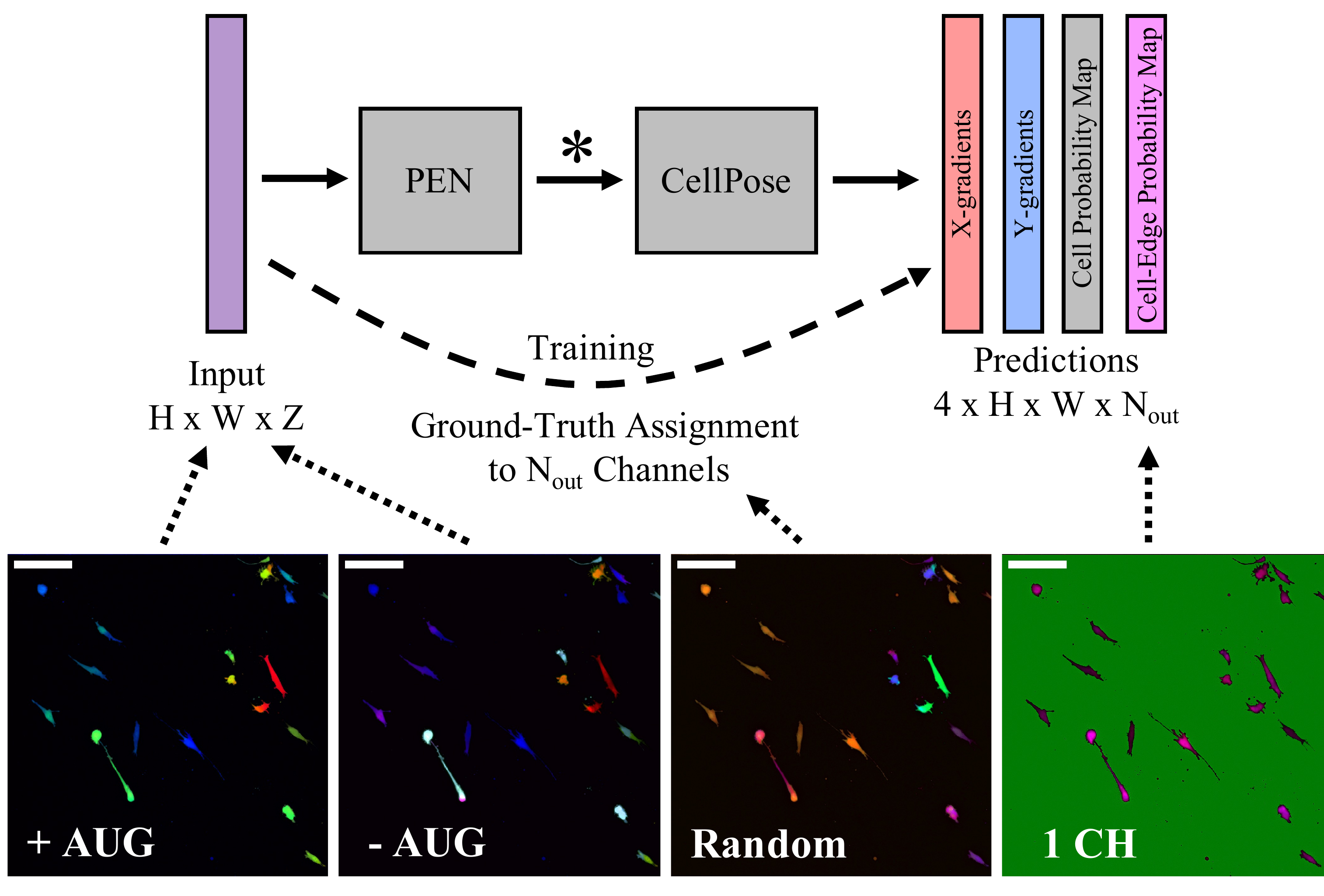}
	    \caption{PEN outputs into CellPose following successful training of networks with different training procedures. At the top, a schematic flow of data during training and inference of PEN+CellPose. During training, multi-channeled ($N_{out}$) ground-truth elements for the outputs of CellPose are used to train the PEN+CellPose network (dotted arrow) from predictions (solid arrows). The images shown at the bottom are all from the output of PEN, indicated by the asterisk on the flow diagram, after training with different training procedures indicated by the dotted arrows pointing to where the alteration is made during training.  (Left to Right) Training with augmentation (+ AUG) and without augmentation (- AUG) using K-means ground-truth assignment ($N_{out}=3$), ground-truth assignment to a single channel (1 CH, $N_{out}=1$), and ground-truth random assignment (Random, $N_{out}=3$).  Scale-bars = 100 $\mu$m.}
	    \label{fig:sup3}
    \end{figure}

    We explore the effects of augmentation on the ability of PEN to encode axial information in its output. As shown in Figure \ref{fig:sup3}, without augmentation, several cells in the image are activated along multiple color channels, in contrast to the depth encoded image learned with augmentation. Particularly, without augmentation, the network struggles to distinguish the axial location of cells that are not at the polar ends of the image stack. We therefore conclude that the augmentations to increase cell density during training are critical to permit the module to learn to encode object depth at intermediate ranges.
    
    The original 2D CellPose network by \cite{stringer2021cellpose} was modified as described in Supplementary \ref{subsec:Supp/CellPoseMods}. The ground truth assignment for cell objects in the training images of CellPose models were assigned as follows. During augmentation, cell centroids from the original image stack are similarly recalculated as augmentations are applied. After, a K-means algorithm utilizing only the Z-positions of cells in the image as features are used to cluster cells into $N_{out}$ distinct groups. The initial conditions have the first cluster set at the lowest cell position, the final cluster set at the highest cell position, and the remaining clusters spaced linearly equidistant from each other. The ground-truth horizontal and vertical flows and cell edges are calculated on the fly from the final stack of cell masks following augmentations, producing identically sized stacks of gradients and cell edges with the axial dimension of size equal to the number of ground truth labels. Finally, the ground-truth elements are assigned to the available $N_{out}$ number of slices corresponding to their K-means assignment.
    
    We investigate the effect of the additional output channels and the assignment strategy on the output of PEN. As shown in \ref{fig:sup3}, removal of the additional output channels by reducing $N_{out}$ to 1, congruent with the original CellPose output, we find that PEN only acts to separate background from object. Furthermore, by randomly assigning ground-truth objects to $N_{out}=3$ output channels as shown in Figure \ref{fig:sup3}, we find that PEN does not learn to embed objects by depth, and greatly reduces the segmentation performance of the network, as shown in Table 2 of the main text. Therefore, we conclude that allowing each output element of the network to have more than a single channel, and an assignment strategy of ground-truth objects to those elements based on position, is essential for the PEN module to learn to encode depth.
    
    In the 2D Mask-RCNN algorithm, we assume that if axial information could be passed to the RPN to detect overlapping instances of cells, then PEN+Mask-RCNN would implement such information during training. Since the Mask-RCNN implementation used in this work \cite{matterport_maskrcnn_2017} needs only the stack of 2D individual masks as ground-truth elements, where each slice has the same resolution as the 2D input image, we pass the 3D image stack as input to PEN and the final stack of cell masks as the ground-truth for the output of Mask-RCNN, following augmentation as previously described.
    

\bibliographystyle{abbrv}
\bibliography{Supplementary}